\documentclass[runningheads]{llncs}

\usepackage{filecontents,lipsum}
\usepackage[noadjust]{cite}
\usepackage{graphics}
\usepackage[pdftex]{graphicx}
\usepackage{mathtools}
\usepackage{subfig}
\usepackage{wrapfig}
\usepackage[left=1in,top=1in,right=1in,bottom=1in]{geometry}

\begin{document}

\title{A Comparison of Various Approaches to Reinforcement Learning Algorithms for Multi-robot Box Pushing}

\titlerunning{A Comparison of Various Approaches to Reinforcement Learning Algorithms}

\author{Mehdi Rahimi\inst{1}, Spencer Gibb\inst{2}, Yantao Shen\inst{1}, and Hung Manh La\inst{2}}

\authorrunning{M. Rahimi et al.}

\institute{Department of Electrical and
Biomedical Engineering
University of Nevada, Reno
Reno, Nevada 89557, USA \and
Advanced Robotics and Automation (ARA) Lab, Department of Computer Science and Engineering 
University of Nevada, Reno
Reno, Nevada 89557, USA. Corresponding email:
        \email{hla@unr.edu}}

\maketitle 

\begin{abstract}
In this paper, a comparison of reinforcement learning algorithms and their performance on a robot box pushing task is provided. The robot box pushing problem is structured as both a single-agent problem and also a multi-agent problem. A Q-learning algorithm is applied to the single-agent box pushing problem, and three different Q-learning algorithms are applied to the multi-agent box pushing problem. Both sets of algorithms are applied on a dynamic environment that is comprised of static objects, a static goal location, a dynamic box location, and dynamic agent positions. A simulation environment is developed to test the four algorithms, and their performance is compared through graphical explanations of test results. The comparison shows that the newly applied reinforcement algorithm out-performs the previously applied algorithms on the robot box pushing problem in a dynamic environment. 

\end{abstract}

\vspace{-10pt}
\section{Introduction}
The robot box pushing problem has been studied in both single and multi-agent cases \cite{Wang06}. The problem entails one or more robotic agents working together to push a box from an arbitrarily starting location to the goal location, through an environment that is often filled with obstacles that the box and robotic agents cannot pass through and should instead avoid. Robotic agents cooperatively pushing an object through an obstacle filled environment has applications in warehouse automation and disaster relief. In the case of the box pushing problem, the benefit of multi-agent collaboration is two-fold: the box can sometimes be too heavy for a single robot to push by itself, and the overall number of actions performed by a robot team to move the box to the goal location can be greatly reduced through collaboration and joint-actions. 

Previous works on the robot box pushing problem can effectively be split into two categories: machine learning solutions and non-machine learning solutions. Of the non-machine learning solutions, there have been several attempts to perform box pushing. Genetic algorithms (GA) have been applied to attempt a Pareto optimal solution that minimizes time taken and energy expended to push a box \cite{Chakraborty09}. GA-based methods did not attempt to include obstacles in the environment and are not necessarily suitable for obstacle ridden environments. The noisy non-dominated sorting bee colony algorithm was applied to a box pushing problem with few obstacles and agents, and focused more on dealing with a noisy environment than the overall box pushing problem \cite{Rakshit16}. It can also be noted that the results of the Q-learning algorithm are affected by uncooperative actions in a way that is similar to noise. Research has also focused on path planning for box pushing given complete knowledge of the environment by modeling objects in the environment as convex polygons and planning the box trajectory \cite{Parra09}. Having complete knowledge of the environment initially is not a realistic approach in most applications, since the position of each robotic agent can affect the environment, so assuming the environment is dynamic is preferred. 

Machine learning solutions to the robot box pushing problem consist primarily of reinforcement learning approaches; specifically Q-learning approaches. Single-agent Q-learning has been extended to allow three agents to push a randomly placed box to the goal location, but no obstacles are presented \cite{Wang05}. Q-learning has been applied to the box pushing problem using decision trees for adaptive state aggregation, but the results show that the algorithm performance is not a significant improvement over single-agent Q-learning, and the environment only contains a single obstacle, making the test case trivial \cite{Hwang14}. Bayesian-discrimination-function-based reinforcement learning (BRL) is applied to allow multiple agents to push a box across a test area, but the area contains no obstacles \cite{Yasuda13}. In \cite{Wang06}, single-agent and multi-agent Q-learning are applied to the box pushing problem in an obstacle ridden environment. The results show, counter-intuitively, that multi-agent Q-learning is not capable of reliably reaching the goal and that the agents cannot learn from their experience, while single-agent Q-learning is reliable and learns. Overall, machine learning solutions out-perform non-machine learning solutions, but results tend to show trivial cases or fail to address aspects of the box pushing problem. 

In this paper, the implementation and results in \cite{Wang06} are extended by applying an additional Q-learning algorithm to the box pushing problem, restructuring the action set for the agents, and changing the reward representation. For comparison, all of the algorithms tested in \cite{Wang06} are also tested. The results show that the extensions to the original method allow the multi-agent team to learn and converge to a policy quicker than the previously implemented algorithms, while also showing an improvement over the single-agent case. 

The remainder of the paper is structured as follows. In Section \ref{sec-2}, a description of the Q-learning algorithms implemented for this problem is provided, including: state-space, actions, and reward function. In Section \ref{sec-3}, the experimental environment is described and experimental results are provided, along with a discussion of the results. Finally, in Section \ref{sec-4}, a conclusion is provided. 

\vspace{-10pt}
\section{Methodology}
\label{sec-2}
Four Q-learning algorithms are implemented in this paper: single-agent Q-learning, multi-agent Q-learning with separate Q-tables for each agent, multi-agent Q-learning with a shared Q-table, and cooperative Q-learning with more frequent updates. In this section, these algorithms will be described, along with a detailed description of the state space representation, the possible actions for each agent, and the method for assigning reward to the agents. 

In this paper, the Q-learning algorithms \cite{La_ICIAE13} are modeled as Markov decision processes (MDPs). An MDP is defined as a tuple; $\big \langle S, A, R, \beta \big \rangle$, where $S$ is a finite set of states, $A$ is a finite set of actions, $R$ is the reward function, and $\beta$ is the discount factor. 

\vspace{-10pt}
\subsection{Reinforcement Learning Algorithms}
\label{RL-algo}
The single agent Q-learning algorithm is given by
\begin{equation}
	Q(s, a)=Q(s, a) + \alpha[r+\beta\max_{a'\in A} Q(s', a')-Q(s, a)]
\end{equation}

where $s$ is the current state, $a$ is the selected action, $\alpha$ is the learning rate, $s'$ is the next state based on the transition function and the selected action, and $a'$ is the next action from state $s'$.

The multi-agent (independent) Q-learning algorithm where each agent has its own Q-table is given by
\begin{equation}
	Q_{i}(s_i, a_i)=(1-\alpha)Q_{i}(s_i, a_i)+ \alpha[r+\beta\max_{a'_i\in A} Q_{i}(s'_i, a'_i)]
\end{equation}
where $Q_{i}$ is agent $i$'s Q-table, $s_i$ is the state of agent $i$ where $s_i \in S$, $a_i$ is the action taken by agent $i$ where $a_i \in A$, $s'_i$ is the next state of agent $i$ where $s'_i \in S$, and $a'_i$ is the next action taken by agent $i$ where $a'_i \in A$.

Similarly, the multi-agent (independent) Q-learning algorithm where all agents share a Q-table is given by
\begin{equation}
	Q(s_i, a_i)=(1-\alpha)Q(s_i, a_i)+ \alpha[r+\beta\max_{a'_i\in A} Q(s'_i, a'_i)]
\end{equation}
with the difference being that $Q_{i}$ is now simply $Q$.

Finally, the cooperative Q-learning algorithm \cite{La2015} is given by
\begin{equation}
	Q_{i}(s_i, a_i) = \omega Q_{i}(s_i,a_i)+(1-\omega)\sum_{j=1}^{N_{i}} Q_{j}(s_j,a_j)
\end{equation}
where $\omega$ is the weight determining whether the agent Q-table or the neighboring agent Q-tables have more impact on the update. $N_{i}$ is the number of neighbors of agent $i$ in its detection range, $s_j$ is the state of neighbor $j$ where $s_j \in S$, and $a_j$ is the action of neighbor $j$ where $a_j \in A$. 

\subsection{State Space Representation}
Similar to other research in this area, the state space for the box pushing environment is represented by 13 bits. Bits 0-4 are the goal orientation $\theta$, which is converted to a bit string through the equation
\begin{equation}
	State_{bits(0-4)} = FLOOR(\theta/(360.0/32)).
\end{equation}
where the $FLOOR$ function rounds toward negative infinity.

\begin{wrapfigure}{r}{0.45\textwidth}
		\vspace{-0pt}
		\centering 
		  \includegraphics[width=2.2in, trim=0mm 40mm 106mm 0mm, clip=true]{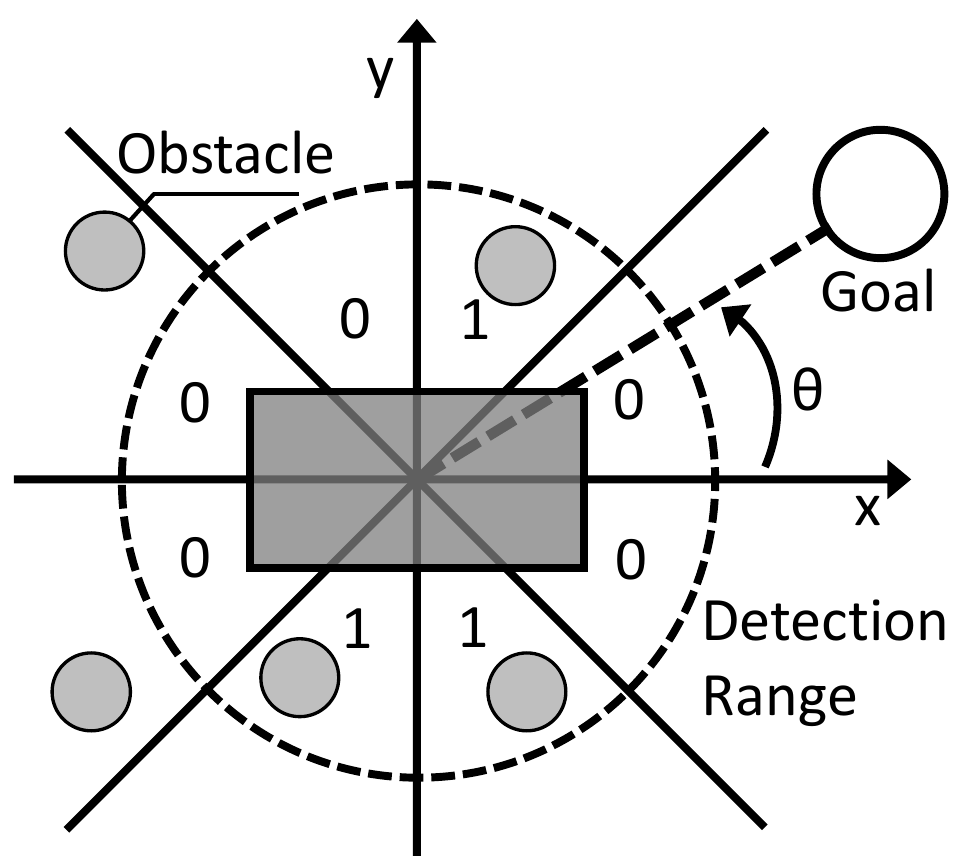}
		\vspace{-0pt}
		\caption{Bits 5-13 of the state space are coded by dividing the surrounding environment into 8 pieces. The presence of an obstacle in any of these pieces (within the detection range) will result in a high bit. The $\theta$ angle is also shown here which represents the relative location of the goal to the box.}
		\label{fig:state}
		\vspace{-20pt}
	\end{wrapfigure}
	
The remainder of the bits are coded based on the presence of obstacles within a respective agent's sensing range. The circle representing the agent's sensing range is split into eight sectors of equal size, with each sector corresponding to a bit in the state representation. If an obstacle is within the sensing range in that sector, then the corresponding bit is a 1, otherwise the bit is a 0. Fig. \ref{fig:state} shows a visual representation of the obstacle coding for the environment. Bits are ordered counter-clockwise starting from zero degrees, and most significant to least significant. In this particular example, the coded state would be $01000110$.

\vspace{-10pt}
\subsection{Action Representation}
\begin{figure}[htp] 
    \centering
    \subfloat[The action space is shown here. Actions 1-4 will only move the box in the pointed directions and actions 5 and 6 will only rotate the box counter clockwise and clockwise respectively.]{%
       \includegraphics[width=2.8in, trim=0mm 25mm 55mm 0mm, clip=true]{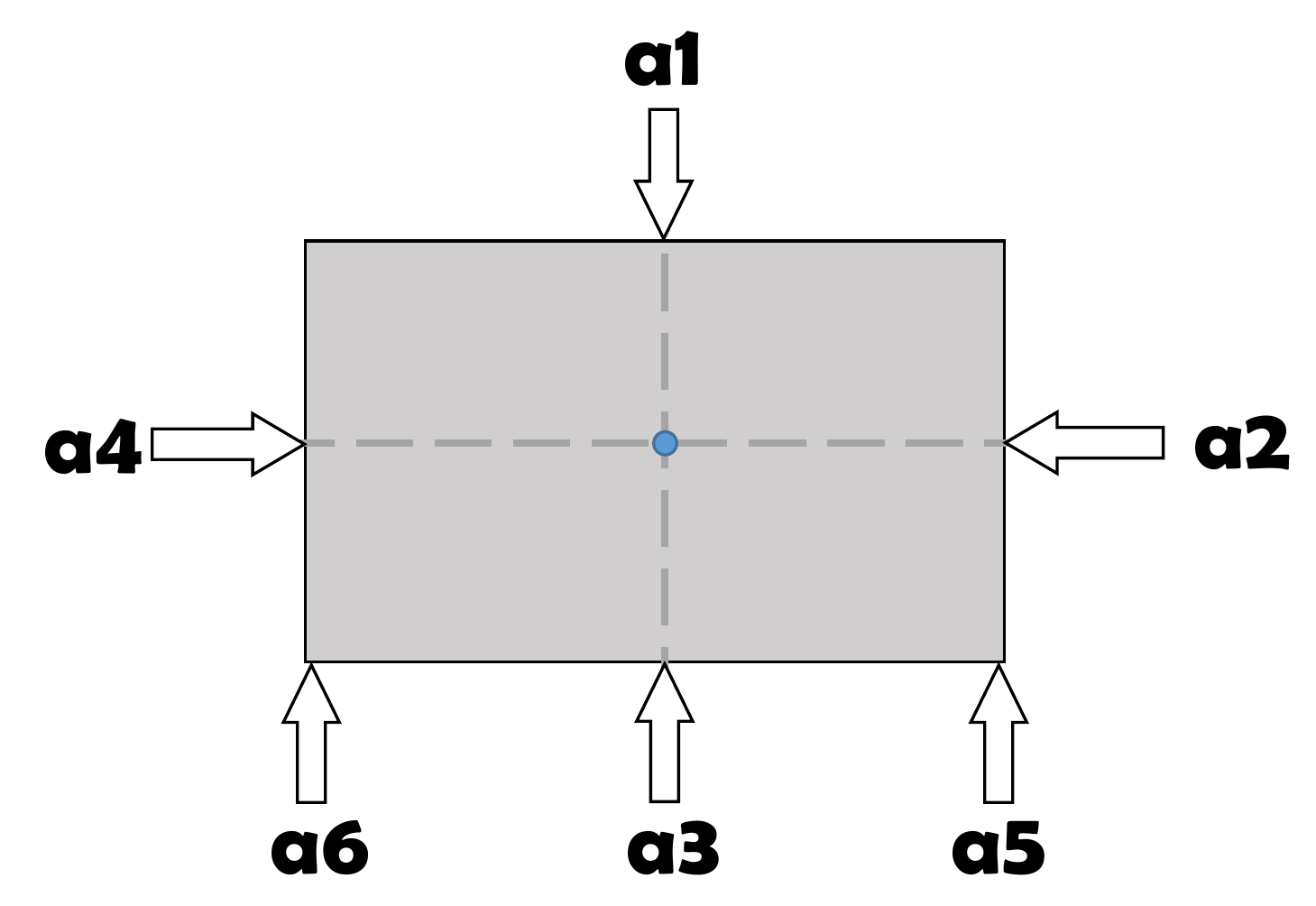}
        \label{fig:actions}%
        }%
    \hfill%
    \subfloat[The new location of the box as a result of the action can be obtained from Eq. \ref{eq:box_movement}.]{%
         \includegraphics[width=3.2in, height = 3.5cm, trim=0mm 85mm 95mm 0mm, clip=true]{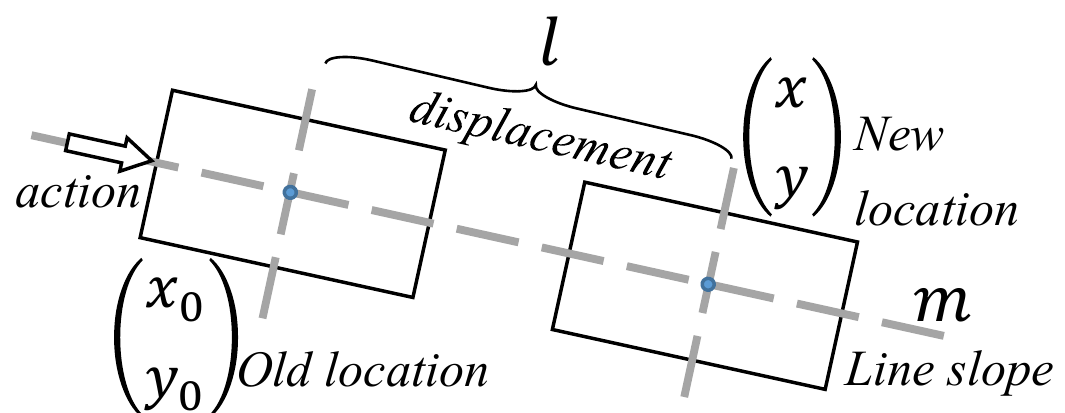}
        \label{fig:box_movement}%
        }%
    \caption{Single-agent Q-learning algorithm and its learning performance.}
    \vspace{-15pt}
\end{figure}

A representation of the actions is shown in Fig. \ref{fig:actions}. As an extension to the previous works in this area, the actions were redefined and simplified as follows: The actions 1 through 4 are only related to the movement of the box whereas actions 5 and 6 will rotate the box. The box is considered heavy enough that the results of the rotation actions will not influence the movement of the box. In other words, the rotation actions will simply rotate the box around its center point without changing the position of the center point. Since the actions can be applied to the box regardless of its orientation, a method needs to be developed to assure the correct movement. To mathematically show the result of the actions 1 through 4, these equations can be considered  where $x$ and $y$ are the new location of the box center after taking either action,  $m$ is the slope of the line that represents the box angle, $l$ is the amount of movement as a result of the action, and $x_0$ and $y_0$ are the old location of the  box center point (see Fig. \ref{fig:box_movement}).
\begin{eqnarray}
x=x_0 \pm l.  \sqrt{\frac{1}{1+m^2}}; 
y=y_0 \pm m.l.\sqrt{\frac{1}{1+m^2}}
\label{eq:box_movement}
\end{eqnarray}

\vspace{-10pt}
\subsection{Reward Function}
Adopted from the previous works \cite{Wang06}, the reward function for all the algorithms implemented in this work are defined by a series of conditions which calculate the total reward. The total reward consists of three parts:

Part 1) The first part of the reward is dedicated to finding the distance between the box and the final goal
\begin{equation}
	R_{distance} = (D_{old} - D_{new}).c_d
\end{equation}

where the $D_{old}$ and $D_{new}$ are measured from the goal to the center point of the box.

Part 2) The second part of the reward will emphasize the rotation of the box 
\begin{equation}
	R_{rotation} = cos(\alpha_2 - \alpha_1)-0.9
\end{equation}

where $\alpha_{1}$ is the previous angle of the box and $\alpha_{2}$ is the new angle of the box. The difference of $\Delta \alpha$ indicates the number of degrees of the rotation of the box. The main benefit of this part is to discourage the rotations of more than $25$ degrees. The objective here to prevent the box from constantly rotating while still providing positive rewards for small rotations that are required to evade obstacles.

Part 3) The final part of the reward is concerned with avoiding the obstacles, $R_{obstacle}$, and is simplified as 
\begin{equation}
  R_{obstacle}=\begin{cases}
    1, & \text{no collision with obstacle}.\\
    -9, & \text{collision with obstacle }.
  \end{cases} 
\end{equation}
 This part of the reward was previously defined as a set of more complicated conditions which in our tests did not show a significant difference.

The total reward is defined as follows
\begin{equation}
	R_{total} = w_1 . R_{distance} + w_2 . R_{rotation} + w_3 . R_{obstacles}
\end{equation}

where $w_{1-3}$ are weights that are assigned manually.

\vspace{-10pt}
\section{Experimental Results}
\label{sec-3}

\subsection{Experimental Environment and the Setup}

A simulation system was developed in MATLAB$^\text{\tiny{\textregistered}}$ to implement the algorithms. A testing environment was defined as an area of $1000 \times 700$. The location of the obstacles was limited to an area of $100-700$ on the x-axis and $100-600$ on the y-axis. The obstacles radius was defined as $10$, and their locations were generated randomly by the simulation program. The locations of the obstacles were kept the same for all the algorithms to give a fair comparison of the performance of the methods. The position of the goal location was defined as $[800, 700]^T$ with a radius of $30$. The box dimensions was set as $120 \times 80$ and in the beginning of each episode, the box was moved to the origin.

The simulation for each algorithm was run with maximum of $2000$ iterations for $80$ episodes. Each episode ends when either the box reaches the goal or the maximum number of iterations is reached.

The following parameters were set for all the algorithms: $\alpha = 0.3$ , $\gamma = 0.4$, $\epsilon = 0.3$, $C_d = 0.9$, $w=0.3$, $w_1=0.7$, $w_2=0.05$, $w_3=0.25$.

\subsection{Single-agent Q-learning Results}

\begin{figure}[htp] 
    \centering
    \subfloat[The final path selected as the result of using the single-agent Q-learning is shown here. It is important to note that although the final path is not the shortest possible one, but it is relatively smart in comparison to multi-agent Q-learning. The 6 obstacles are shown with the red circles, and the goal is represented with the black doughnut-shape.]{%
        \includegraphics[width=0.45\textwidth]{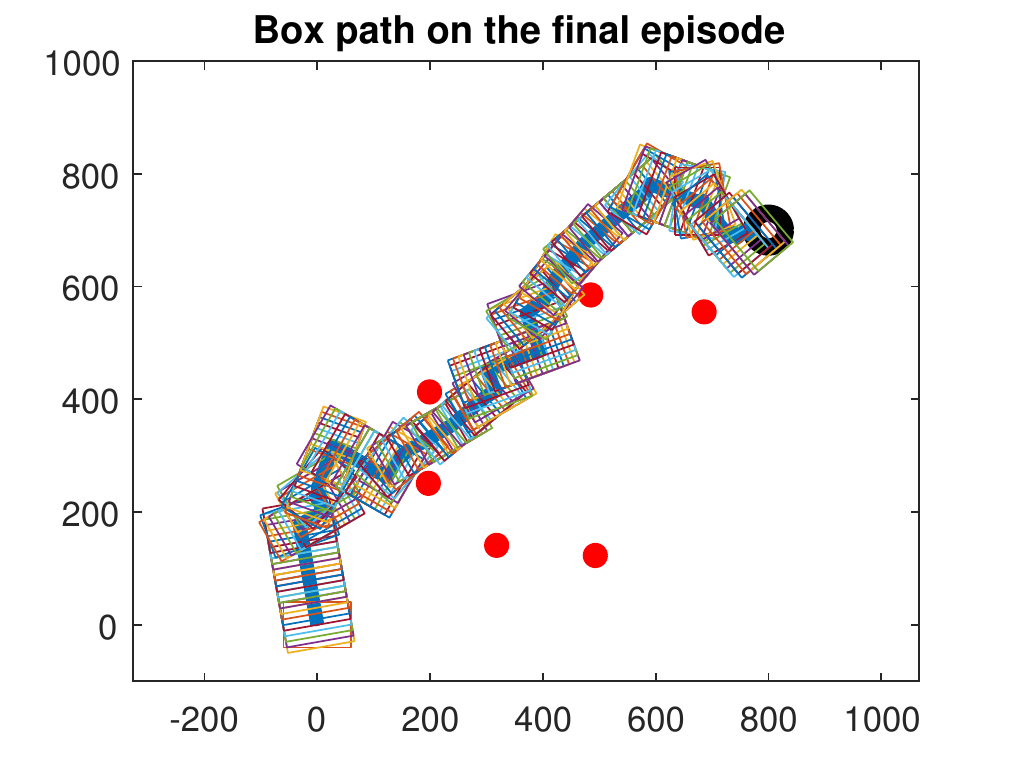}%
        \label{fig:single_path}%
        }%
    \hfill%
    \subfloat[The number of iterations used in each episode before reaching the goal is shown here. It can be seen that the single-agent Q-learning algorithm learns the best path after only 20 episodes.]{%
        \includegraphics[width=0.45\textwidth]{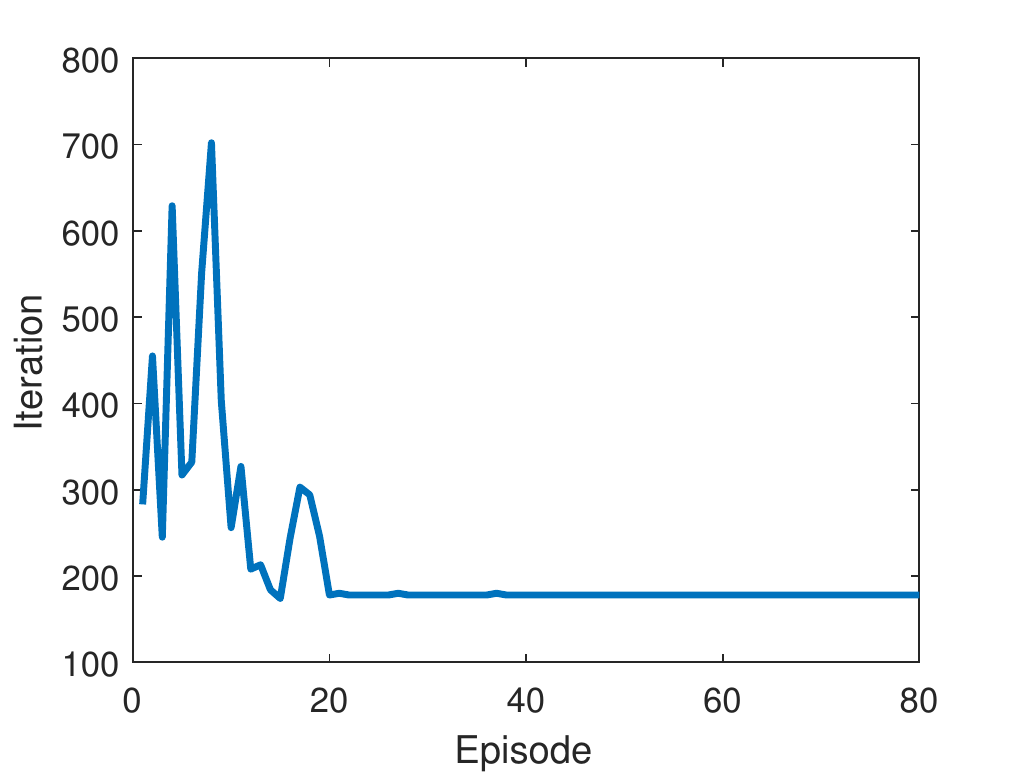}%
        \label{fig:single_ite}%
        }%
    \caption{Single-agent Q-learning algorithm and its learning performance.}
    \vspace{-10pt}
\end{figure}

As described in the methodology section of this paper, the single-agent Q-learning is implemented here, and the result of the final selected path can be seen in Fig. \ref{fig:single_path}.

\begin{figure}[htp] 
    \centering
    \subfloat[The final selected path for the Multi-agent Q-learning Results with Separate Q-tables is shown here. The number of redundant actions is too high as illustrated by the drawings of the box movements.]{%
        \includegraphics[width=0.45\textwidth]{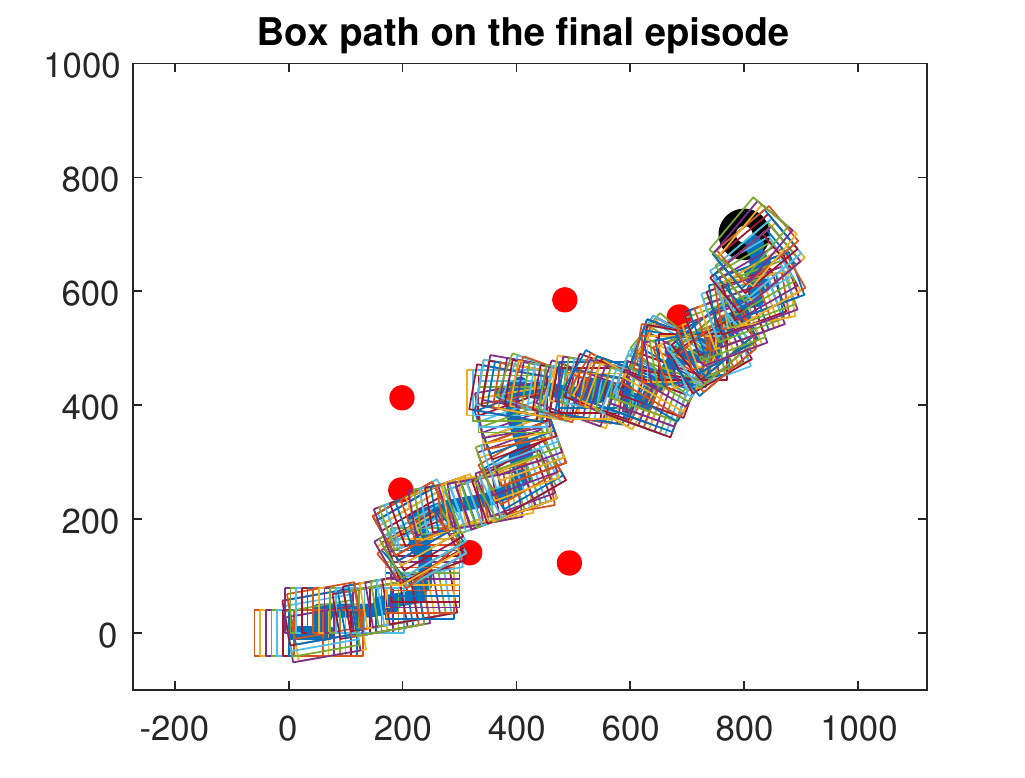}%
        \label{fig:team_separateQ_path}%
        }%
    \hfill%
    \subfloat[The number of iterations used in each episode before reaching the goal is shown here. It can be seen that the multi-agent Q-learning with separate Q-tables does not learn the best path even after 80 episodes although a convergence can be seen.]{%
        \includegraphics[width=0.45\textwidth, height =5.5cm]{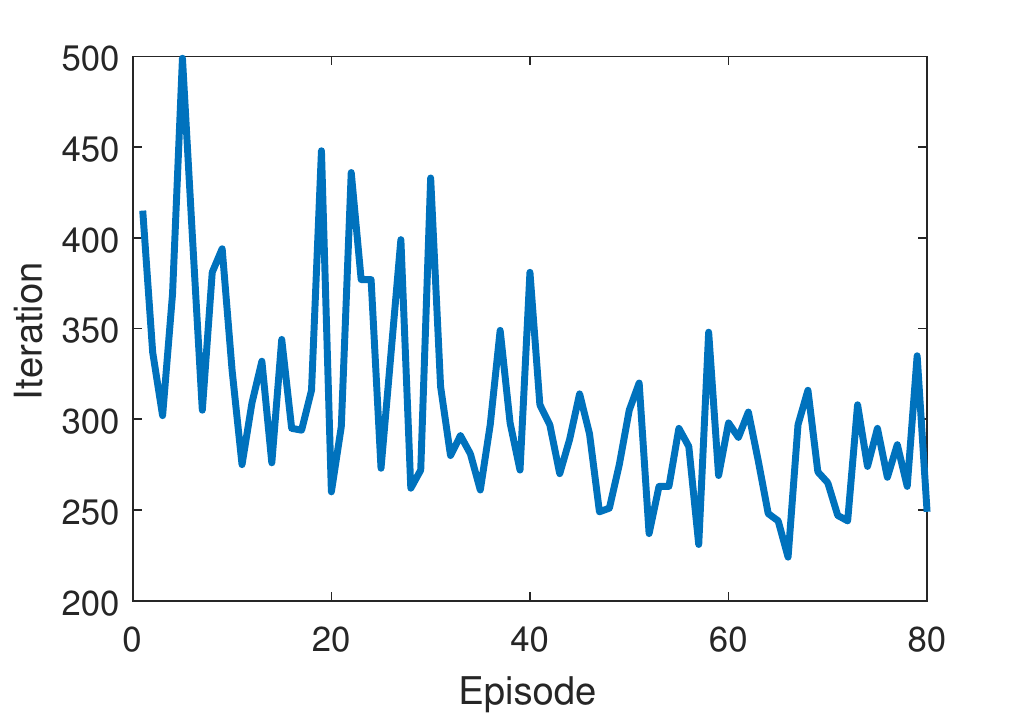}%
        \label{fig:team_separateQ_ite}%
        }%
    \caption{Multi-agent Q-learning with Separate Q-tables algorithm and its learning performance.}
    \vspace{-20pt}
\end{figure}

As it is shown in Fig. \ref{fig:single_path}, the algorithm has managed to choose a path that is very close to a straight line. Although an exact straight line path is not possible in this situation because of the obstacles.

A better representation of the performance of the algorithm can be shown as the number of iterations that were taken in each episode before reaching the goal (see Fig. \ref{fig:single_ite}). This figure shows that in the beginning of the learning process, the algorithm has to try numerous iterations (in this situation, up to 700 at one point) to find a path to reach the goal. But as the learning method progresses, the algorithm finds the shortest path (in this case, after 20 episodes) to reach the goal.

A conclusion can be drawn based on these results that the single-agent Q-learning is effective and can find the best path in a very reasonable amount of time, but a comparison between this and the other algorithms is needed to find whether the single-agent Q-learning is the most effective algorithm. 

\vspace{-10pt}
\subsection{Multi-agent Q-learning Results with Separate Q-tables}

As discussed in the methodology section, this algorithm will use multiple robots (in this case, 3 robots) to perform the task. Each robot will take an optimum action based on the method defined in section \ref{RL-algo} and will update its own Q-table based on the reward. Figure \ref{fig:team_separateQ_path} shows the final selected path using this approach.

Noting Fig. \ref{fig:team_separateQ_path}, it is clear that the the box did reach the goal on the final path but there are many redundant actions that have taken place. It can be argued that the best path was not selected here as the path can be considered as stepped shaped. A better representation of the performance of this algorithm is shown in Fig. \ref{fig:team_separateQ_ite}. It can be seen that the algorithm does not converges to a line even after 80 episodes although the number of iterations is certainly decreasing. This shows that the 80-episode limit was not sufficient here. The number of iterations in each episode is also noticeably high. This is another sign that the algorithm is not the most efficient one.

\vspace{-10pt}
\subsection{Multi-agent Q-learning Results with a Shared Q-table}

\begin{figure}[htp] 
    \centering
    \subfloat[The path selected in the final episode in the Multi-agent Q-learning Results with a Shared Q-table is shown here. Based on the figure, it can be said that the optimum path was not selected after 80 episode.]{%
        \includegraphics[width=0.45\textwidth]{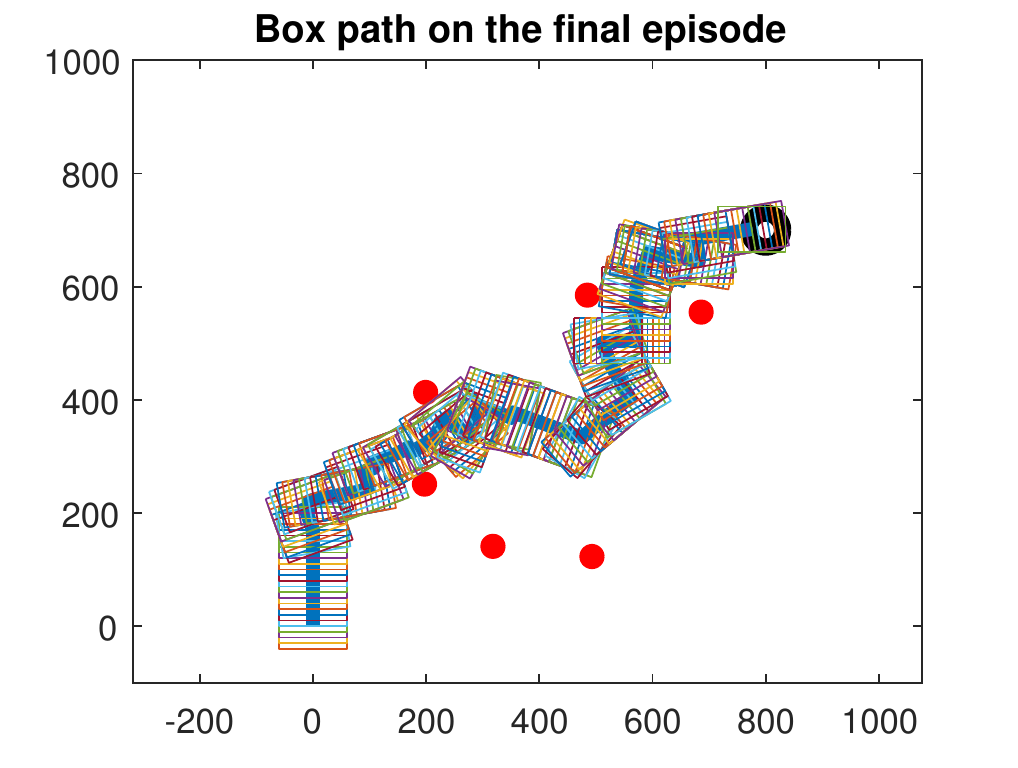}%
        \label{fig:team_sharedQ_path}%
        }%
    \hfill%
    \subfloat[The number of iterations used in each episode before reaching the goal is shown here. It can be seen that the multi-agent Q-learning with a shared Q-table algorithm is converging but the learning process has not stopped even after 80 episodes.]{%
        \includegraphics[width=0.45\textwidth, height =5.5cm]{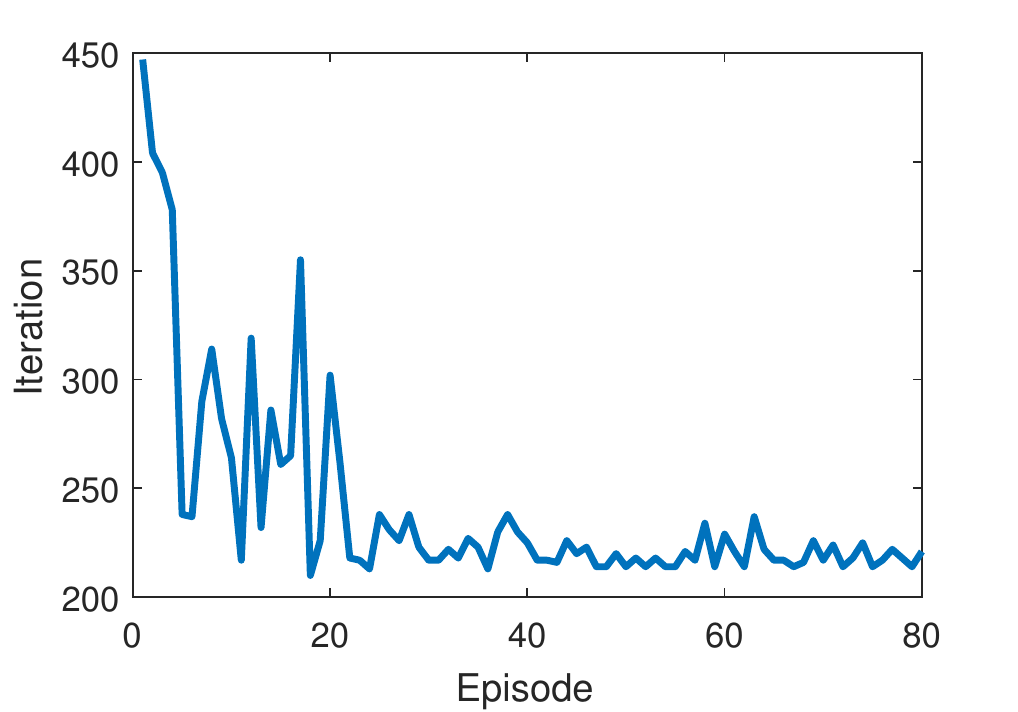}%
        \label{fig:team_sharedQ_ite}%
        }%
    \caption{Multi-agent Q-learning with a Shared Q-table algorithm and its learning performance.}
    \vspace{-10pt}
\end{figure}

The main difference between this algorithm and the previous one is that in this case, only one Q-table is used for all the robots. Each robot takes an optimum action and updates this sole Q-table based on the reward. The Q-table then is shared between all other robots for future action selection. The final selected path for this algorithm is shown in Fig. \ref{fig:team_sharedQ_path}.

Fig. \ref{fig:team_sharedQ_ite} examines the performance of this algorithm. This result shows a significant improvement in comparison to the multi-agent Q-learning with separate Q-tables. A convergence can be seen in the number of required iterations that happened after about 20 episodes. It can be said that using only one shared Q-table can result in significantly better performance than using separate Q-tables for each robot.

It should be noted that although the box does reach to the goal in every episode, this cannot be counted as the only requirement to count an algorithm as efficient. Since the environment dimensions are not extremely large, there is always a factor of chance of getting to the goal after several random movements. Therefore, the efficiency of the algorithm should be assessed on how fast this was achieved.

\vspace{-10pt}
\subsection{Cooperative Q-learning with Frequent Updates}
     \vspace{-15pt}
\begin{figure}[htp] 
    \centering
    \subfloat[The final selected path after 80 episodes of the Cooperative Q-learning with Frequent Updates algorithm is shown here. It can be argued that the cooperative Q-learning with frequent updates algorithm has found the best possible path in this case.]{%
        \includegraphics[width=0.45\textwidth]{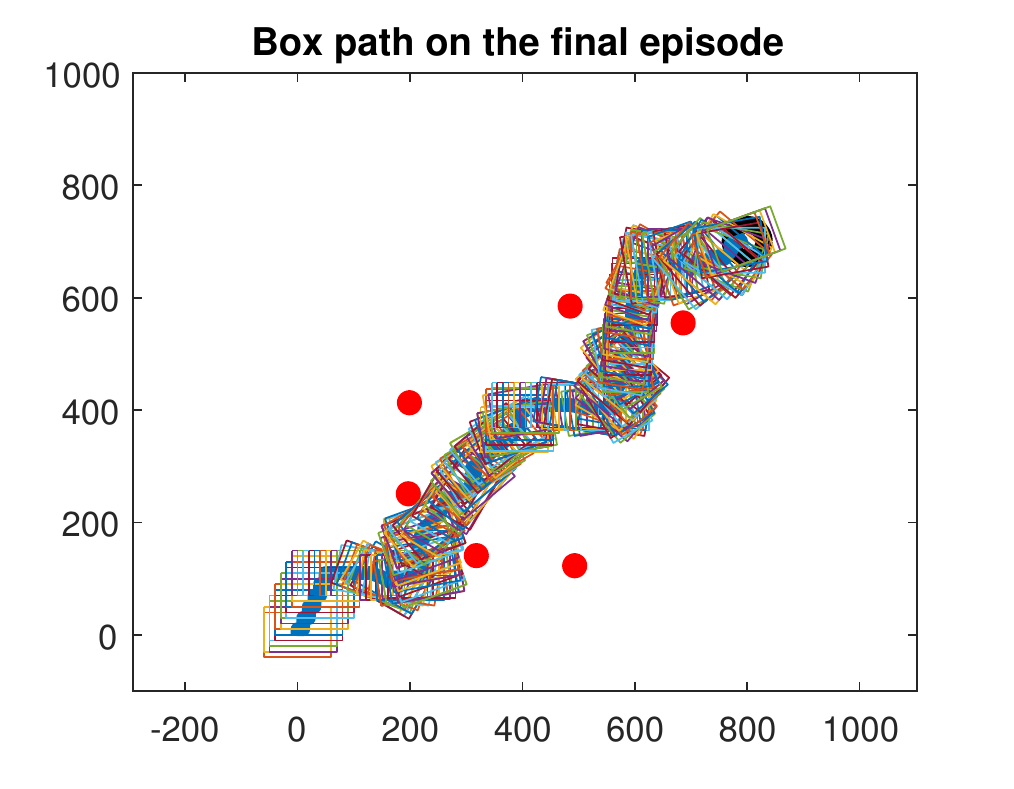}%
        \label{fig:team_ourway_path}%
        }%
    \hfill%
    \subfloat[The number of iterations used in each episode before reaching the goal is shown here. It can be seen that the algorithm manages to reach the goal very fast and converges to the the best path very quickly.]{%
        \includegraphics[width=0.45\textwidth, height =5.5cm]{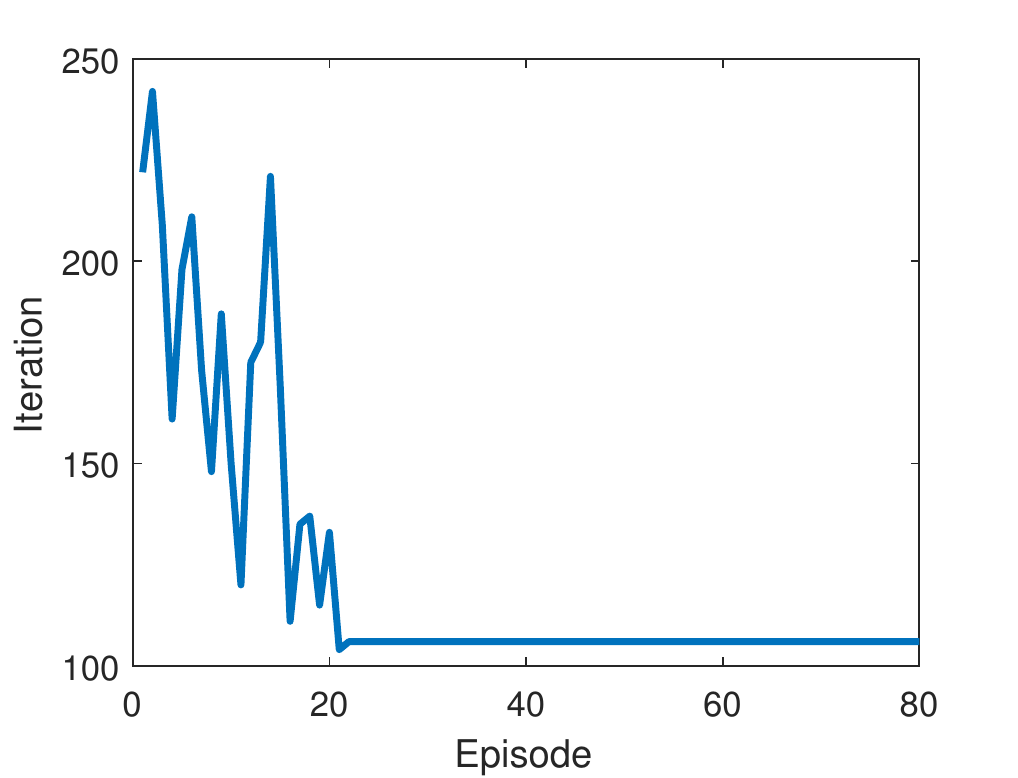}%
        \label{fig:team_ourway_ite}%
        }%
    \caption{Cooperative Q-learning with Frequent Updates algorithm and its learning performance.}
     \vspace{-15pt}
\end{figure}

\begin{wrapfigure}{r}{0.45\textwidth}
  \centering
  \includegraphics[width=3in]{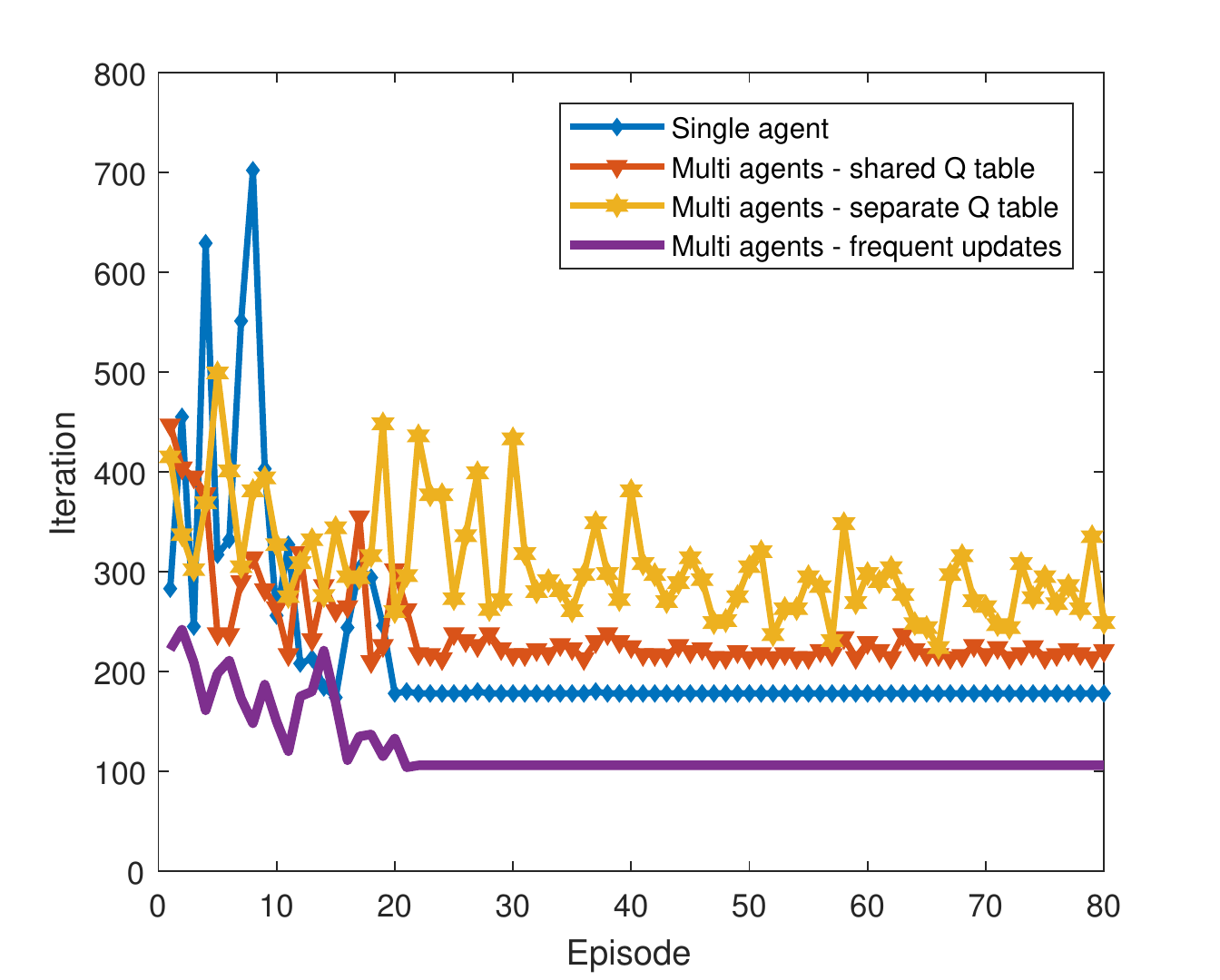}
  \caption{A comparison between the four algorithms is shown here. The Cooperative Q-learning with Frequent Updates (shown in the purple line) is performing significantly better than the other three with noticeably less number of iterations in each episode and converging very quickly.}
  \label{fig:comparison}
  \vspace{-20pt}
\end{wrapfigure}
Finally, the proposed algorithm is implemented. The cooperative algorithm is superior to the multi-agent Q-learning in that each robot uses the results from other robots to improve its action selection.  The "Frequent Updates" refers to the fact that the Q-table is also updated more frequently than either of the multi-agent Q-learning algorithms because each robot updates its Q-table after each action instead of one Q-table update for all the robots as happens in multi-agent Q-learning. The result of this algorithm can be seen in Fig. \ref{fig:team_ourway_path}. The performance of this Cooperative Q-learning can be seen in Fig. \ref{fig:team_ourway_ite}.

The result shown in Fig. \ref{fig:team_ourway_ite} shows a significant improvement over both multi-agent Q-learning algorithms. The algorithm needs significantly less iterations in each episode to reach the goal and converges to the best path after only 20 episodes. The number of iterations needed is also less than half of the same in the multi-agent Q-learning algorithms.

Overall, it can be concluded that the Cooperative Q-learning with Frequent Updates presents significantly better results than the multi-agent Q-learning algorithms (either with separate Q-tables or with a shared Q-table).

\subsection{Comparison Between the Algorithms}

To better discuss the performance of these four algorithms, Fig. \ref{fig:comparison} shows a comparison between them.

It is evident from this figure that the Cooperative Q-learning with Frequent Updates is performing significantly better than the other algorithms. Not only the box reaches the goal in less iterations in each episode, but also the algorithm converges surprisingly fast.

It is noteworthy to mention that based on the comparison figure, the single-agent Q-learning performs better than the multi-agent Q-learning.

\begin{figure}[htp] 
    \centering
    \subfloat[All the paths selected in each episode of the multi-agent algorithm is shown here. Many different paths needed to be examined until the best one is selected.]{%
        \includegraphics[width=0.45\textwidth]{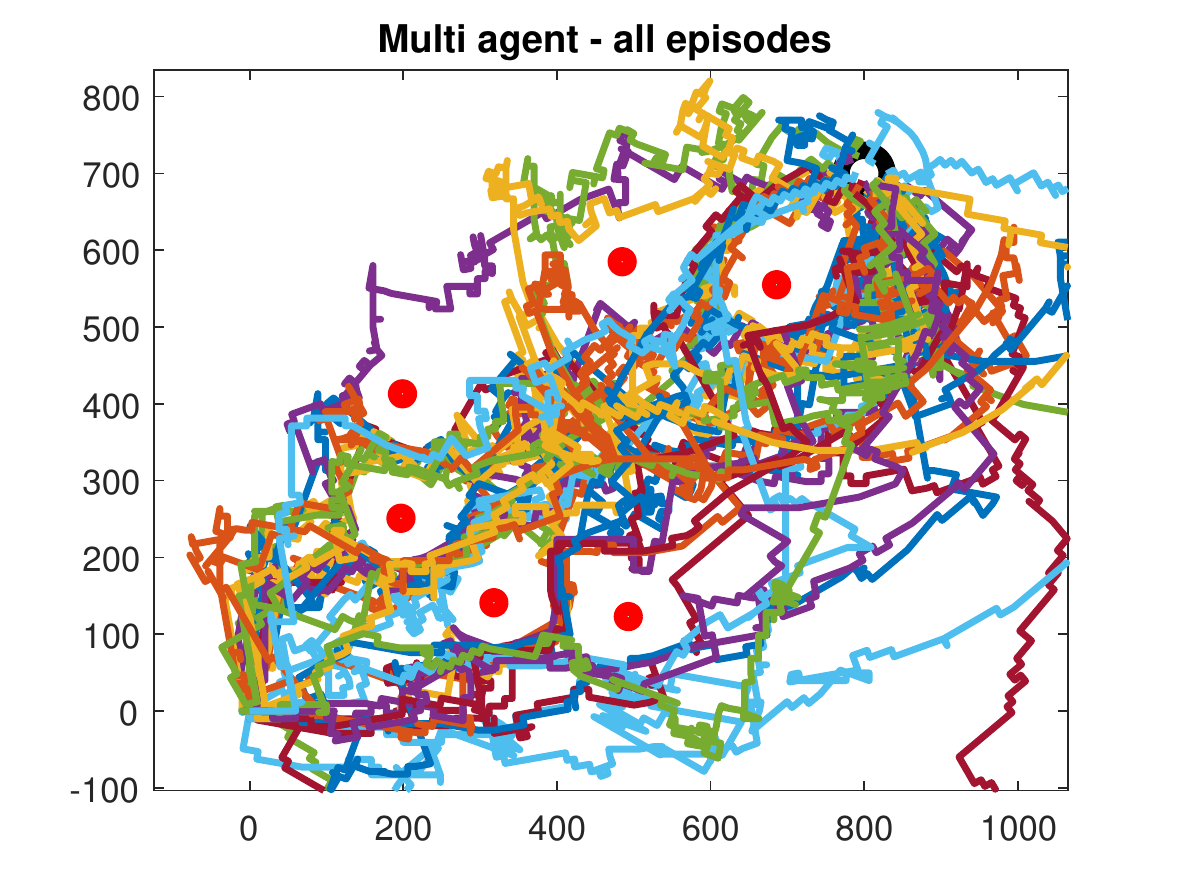}%
        \label{fig:multi_all_eps}%
        }%
    \hfill%
    \subfloat[All the paths selected in each episode of the single-agent algorithm is shown here. The number of examined paths before selecting the final path is significantly less than the multi-agent.]{%
        \includegraphics[width=0.45\textwidth]{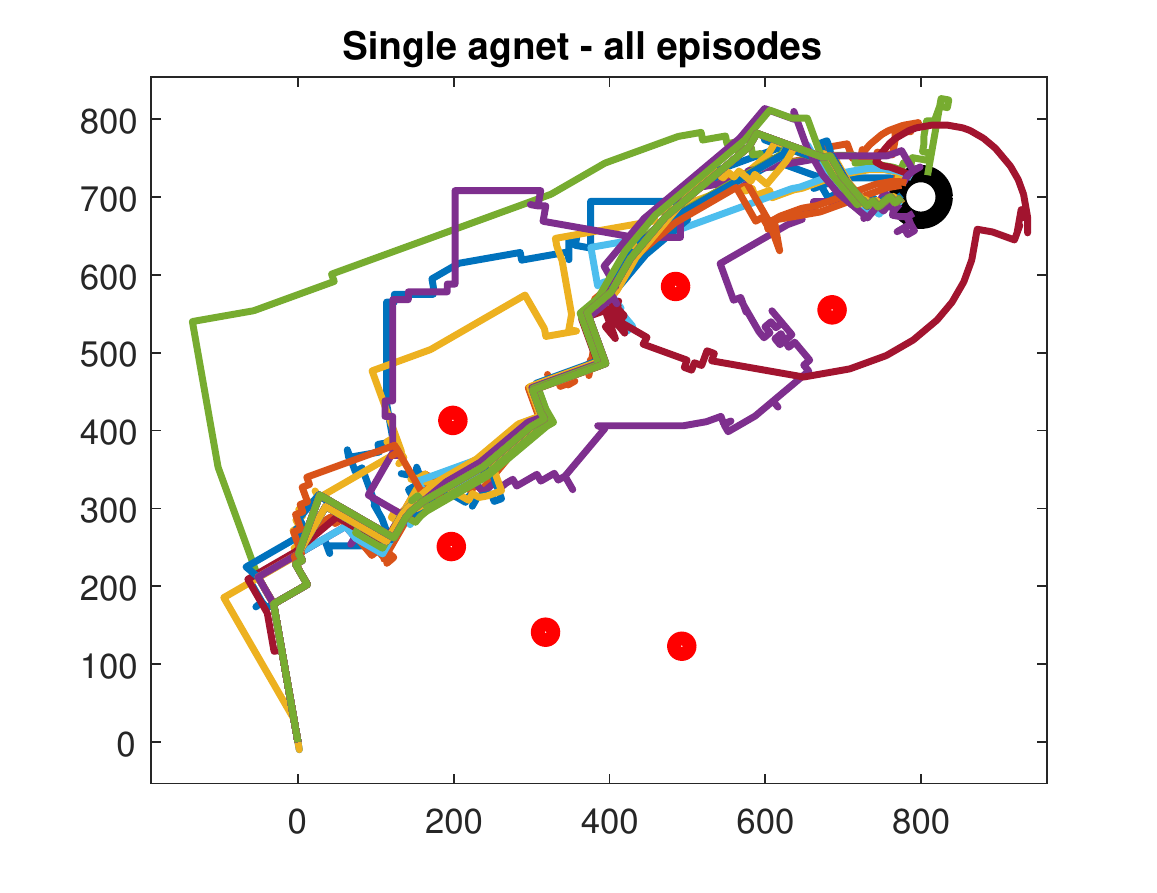}%
        \label{fig:single_all_eps}%
        }%
    \caption{Multi-agent algorithm and single-agent algorithm} 
\vspace{-15pt}
\end{figure}

A comparison between the multi-agent algorithm and the single-agent algorithm was done too. Fig. \ref{fig:multi_all_eps} shows all the paths that were examined in a multi-agent algorithm before selecting the final one.

As a comparison, Fig. \ref{fig:single_all_eps} shows the same result for the single-agent algorithm.

It can be seen that significantly less paths are examined in the single-agent algorithm. The values of the Q-tables are the determinant factor in either case and it is very hard to explain the reason of these behaviors mathematically. 

\section{Conclusion}
\label{sec-4}

This work presents our implementation and comparison of four reinforcement algorithms. These were: single-agent Q-learning, multi-agent Q-learning with separate Q-tables for each agent, multi-agent Q-learning with a shared Q-table, and cooperative Q-learning with more frequent updates.

A framework was implemented in MATLAB to easily compare the performance of each of these algorithms. The results was shown graphically as the final selected path for each algorithm and also the number of required iterations in each episode before reaching the goal. The comparison between the results showed that the cooperative Q-learning with more frequent updates performed significantly better than the other algorithms. In any case, the multi-agent Q-learning algorithms (either with separate Q-tables or a shared Q-table) did not perform as well as the single-agent Q-learning algorithm.

The cooperative Q-learning with more frequent updates was able to reach the goal in significantly less number of iterations and also converged to the best path astonishingly fast.

The overall conclusion can be that the multi-agent Q-learning algorithms should be replaced with the cooperative Q-learning whenever possible. Also, the number of times that the robots update their Q-tables should be more frequent as this resulted in a remarkably better and faster convergence to the best path.

\bibliography{paperCS}
\bibliographystyle{splncs04}

\end{document}